\documentclass{article}

\usepackage[preprint]{colm2025_conference}

\usepackage{microtype}
\usepackage{hyperref}
\usepackage{url}
\usepackage{booktabs}
\usepackage{tabularx} 
\usepackage{lineno}

\definecolor{darkblue}{rgb}{0, 0, 0.5}
\hypersetup{colorlinks=true, citecolor=darkblue, linkcolor=darkblue, urlcolor=darkblue}

\usepackage{inconsolata}

\usepackage{soul}
\usepackage{url}
\usepackage[utf8]{inputenc}
\usepackage[small]{caption}
\usepackage{graphicx}
\usepackage{amsmath}
\usepackage{amsthm}
\usepackage{enumitem}
\usepackage{booktabs}
\usepackage{algorithm}
\usepackage{algorithmic}
\usepackage{amsfonts}       
\usepackage{nicefrac}       
\usepackage{microtype}      
\usepackage{xcolor}         
\usepackage{amsmath}         
\usepackage{makecell}
\usepackage{siunitx}
\usepackage{amssymb} 
\usepackage[utf8]{inputenc}
\usepackage{caption}  
\usepackage{subfigure}
\usepackage{float}
\usepackage{subcaption}
\usepackage{stfloats}
\usepackage{CJKutf8}
\usepackage{amsmath,amssymb}
\usepackage{amsthm}
\newtheorem{theorem}{Theorem}[section]
\newtheorem{lemma}[theorem]{Lemma}
\usepackage{microtype}
\usepackage{hyperref}
\usepackage{booktabs}
\usepackage{multirow}
\usepackage{tcolorbox}
\usepackage{cleveref}

\tcbuselibrary{breakable}
\tcbset{
  breakable,
  break at={\textheight-2\baselineskip},
  break automata/.code={\bottomrule},
  before={\par\noindent\nobreak}
}

\newtcolorbox{mybox}{
  colback=red!10, 
  colframe=red!75!black, 
  boxrule=1pt, 
  arc=4pt, 
  boxsep=0pt, 
  left=6pt, 
  right=6pt, 
  top=6pt, 
  bottom=6pt, 
  width=\textwidth 
}

\newtcolorbox{mybox1}{
  colback=gray!10, 
  colframe=gray!75!black, 
  boxrule=1pt, 
  arc=4pt, 
  boxsep=0pt, 
  left=6pt, 
  right=6pt, 
  top=6pt, 
  bottom=6pt, 
  width=\textwidth 
}

\usepackage{xcolor}
\usepackage{soul}

\definecolor{myblue}{rgb}{0.651, 0.808, 0.890}
\definecolor{mygreen}{rgb}{0.698, 0.875, 0.541}
\definecolor{myorange}{rgb}{0.992, 0.749, 0.435}

\title{Enhancing Commentary Strategies for Imperfect Information Card Games: A Study of Large Language Models in Guandan Commentary}

\author{
    Author Name
    \affiliations
    Affiliation
    \emails
    email@example.com
}

\author{
  \begin{tabular}{c}
    Meiling Tao $^{1}$\thanks{$^{}$Equal contribution.}, 
    Xuechen Liang $^{2*}$, 
    Xinyuan Song $^{3}$, 
    Yangfan He $^{4}$, 
    Yiling Tao $^{5}$, \\
    Jianhui Wang $^{6}$, 
    Sun Li $^{7}$, 
    Tianyu Shi $^{8}$\thanks{$^{*}$Corresponding author: \texttt{ty.shi@mail.utoronto.ca}} \\[1ex]
    $^1$ Guangdong University of Technology,
    $^2$ East China Jiaotong University, \\
    $^3$ Emory University,
    $^4$ University of Minnesota - Twin Cities, \\
    $^5$ South China University of Technology, \\
    $^6$ University of Electronic Science and Technology of China, \\
    $^7$ Amazon,
    $^8$ University of Toronto \\ 
  \end{tabular}
}

\begin{document}

\maketitle

\begin{abstract}

Recent advancements in large language models (LLMs) have unlocked the potential for generating high-quality game commentary. However, producing insightful and engaging commentary for complex games with incomplete information remains a significant challenge. In this paper, we introduce a novel commentary method that combine Reinforcement Learning (RL) and LLMs, tailored specifically for the Chinese card game \textit{Guandan}. Our system leverages RL to generate intricate card-playing scenarios and employs LLMs to generate corresponding commentary text, effectively emulating the strategic analysis and narrative prowess of professional commentators. The framework comprises a state commentary guide, a Theory of Mind (ToM)-based strategy analyzer, and a style retrieval module, which seamlessly collaborate to deliver detailed and context-relevant game commentary in the Chinese language environment. We empower LLMs with ToM capabilities and refine both retrieval and information filtering mechanisms. This facilitates the generation of personalized commentary content. Our experimental results showcase the substantial enhancement in performance achieved by the proposed commentary framework when applied to open-source LLMs, surpassing the performance of GPT-4 across multiple evaluation metrics. Our code is available at \url{https://github.com/heimy2000/guandan}.

\end{abstract}

\vspace{-0.5em}
\section{Introduction}
\vspace{-0.5em}

Large language models (LLMs) make significant advances in the field of natural language processing in recent years, particularly demonstrating exceptional capabilities in complex text generation and context understanding. This opens up new possibilities in the field of game commentary generation, enabling the production of insights-rich and contextually detailed commentary. However, generating high-quality game commentary still faces multiple challenges, including a deep understanding of game rules and strategies, the smooth generation of real-time commentary, and engaging narrative skills.

Previous work makes some progress in game commentary generation. For example, ~\cite{liao1990computer,sadikov2007automated} conduct preliminary explorations into chess commentary, but these methods are mainly based on simple rules and limited by the scale and quality of datasets. ~\cite{kameko2015learning} tries advanced techniques in Shogi games, generating commentary by extracting key terms from game states and integrating them with language models, marking a step towards more complex generation methods. However, game commentary generation still faces a series of challenges, especially in handling incomplete information games~\cite{Suspicion-Agent}, simulating human commentators' advanced strategy analysis, and applications in non-English environments. Moreover, existing methods still fall short in data-driven deep learning applications, limiting the naturalness and richness of generated commentary~\cite{kim2020automatic,taniguchi2019generating}.

\begin{figure*}[!ht]
  \centering
  \includegraphics[width=1\textwidth]{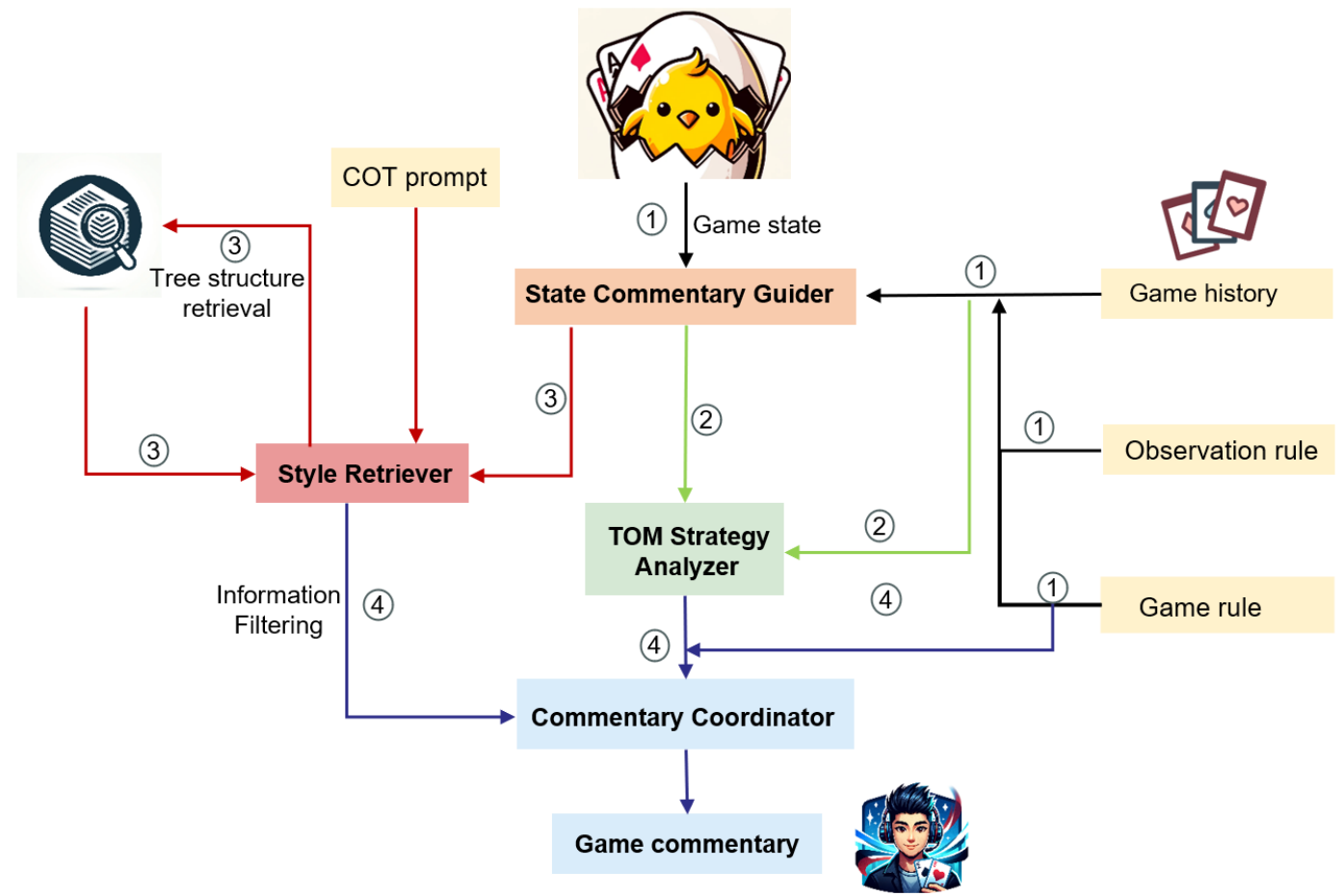}
  \captionof{figure}{
  It illustrates the commentary process of a Guandan game, with inputs including the game state, game history, and the corresponding game rules and observation rules. (1) The system first uses a State Commentary Guider to transform these inputs into preliminary commentary text. (2) The ToM Strategy Analyzer receives this text and utilizes ToM to analyze players’ strategies and behaviors, predicting opponents' potential psychological states and reactions.(3) A Style Retriever using COT prompts employs a tree-based retrieval method and information filtering system to extract statements that match a specific commentary style. (4) The Commentary Coordinator integrates all the commentary text to produce the final game commentary.
  }

  \label{fig:1}
\end{figure*}

To assess the performance of LLMs in cooperative and incomplete information environments, we select the popular Chinese card game Guandan\footnote{https://en.wikipedia.org/wiki/Guandan}. The game's objective is for two players on the same team to play out all their cards as quickly as possible to score higher team points. Consistent with recent work, we evaluate commercial and Chinese open-source models in a zero-training setup on the state-of-the-art RL agent Danzero+'s~\cite{zhao2023danzero+} playing behavior to assess their ability to understand and interpret agent actions in a cooperative, incomplete information environment.

In this study, we develop a novel commentary method that combines RL and LLM, specifically designed for the complex, incomplete information card game Guandan. By utilizing RL to generate complex card-playing scenarios, and LLM to generate corresponding commentary text, our system effectively simulates the strategy analysis and narrative skills of professional commentators. The system consists of several core components, including a state commentary guide, a ToM-based strategy analyzer~\cite{DBLP:journals/corr/abs-2302-02083}, and a style retrieval module, as illustrated in Figure~\ref{fig:1}. These modules work together to provide detailed and context-relevant game commentary, while optimizing commentary effects in the Chinese environment. Experimental results show that our commentary framework significantly enhances the performance of open-source LLMs, significantly outperforming GPT-4 on multiple evaluation metrics.

\vspace{-0.5em}
\begin{itemize}[leftmargin=10pt]
    \item 
    We propose a novel commentary method, combining RL and LLM, and extensively evaluate the commentary performance in the complex card game Guandan. 
    \item 
    We endow LLMs with ToM capabilities and optimize retrieval and information filtering mechanisms, allowing LLMs to effectively commentate on the game-playing processes of RL agents and generate more personalized commentary content.
   \item
    We develop a commentary agent framework specifically designed for Guandan, a card game with incomplete information. The commentary agent is capable of effectively handling and explaining complex game situations without the need for special training in the gaming environment.
    \item 
    Our experimental results show that the commentary framework significantly improves the commentary effects of open-source LLMs, far surpassing GPT-4 on multiple metrics.
\end{itemize}

\vspace{-1em}
\section{Related work}
\vspace{-0.5em}
\vspace{-0.5em}
\subsection{Game Commentary Generation}
\vspace{-0.5em}
In the field of game commentary generation, research makes significant progress in recent years. Initially, due to the lack of large-scale training data, researchers primarily use rule-based methods, such as generating commentary through rules in sports and chess games to enhance the viewer experience ~\cite{liao1990computer,sadikov2007automated}. With the development of deep learning technologies, neural networks begin to be applied to commentary generation, such as detailed commentary for soccer and baseball games~\cite{puduppully2021data,gardent2017creating}, and using encoder-decoder models~\cite{taniguchi2019generating} and hierarchical models to generate commentary for eSports~\cite{wang2022esports} games like League of Legends. Additionally, the rise of LLMs brings a new dimension to game commentary, handling multiple tasks through zero-shot learning~\cite{wei2021finetuned} and providing a superior user experience.

Researchers also explore methods combining visual and structured data, such as in racing games where~\cite{ishigaki2021generating} generate automatic commentary by combining multiple data sources. Meanwhile,~\cite{jhamtani2018learning} introduce a large-scale commentary dataset in the field of chess, improving the accuracy and fluency of the generated commentary. More recently,~\cite{nimpattanavong2023fighting} apply LLMs to the generation of commentary for fighting games, demonstrating the model's potential in generating diverse and preferential commentary. Based on these research findings, this paper aims to further explore the application of methods combining reinforcement learning and LLM in Guandan, a complex card game with incomplete information, to further advance the field of game commentary generation.

\vspace{-0.5em}
\subsection{Retrieval-Augmented Generation}
\vspace{-0.5em}
In the field of Retrieval-Augmented Generation (RAG), significant progress has been made recently. One study~\cite{Shahul2023RAGAS:} introduces the RAGAs evaluation framework for rapid assessment of Retrieval-Augmented Generation system performance, particularly suitable for the rapid deployment of LLMs.

Another study investigates the domain adaptability of RAG models and proposes the RAG-end2end approach, enabling them to adapt to specific domain knowledge bases~\cite{Siriwardhana2022Improving}. Additionally, researchers propose MuRAG, a multi-modal Retrieval-Augmented Generator. It leverages external non-parametric multi-modal memory for enhanced language generation, demonstrating outstanding cross-dataset performance in tasks involving image and text-based questions and answers~\cite{Chen2022MuRAG:}. Finally, another study develops the ARM-RAG system, which enhances problem-solving performance by storing and retrieving inference chains. This effectively boosts the intelligence of large language models while reducing training costs~\cite{Melz2023Enhancing}. These advancements demonstrate the potential of RAG technology in enhancing the accuracy of knowledge access, the quality of generation, and the narrative ability. Our research explores the integration of RAG with domain-specific strategies to enhance the strategic depth of game commentary.
\vspace{-0.5em}
\section{Method}
\vspace{-0.5em}
The study introduces a modular approach (Figure~\ref{fig:1}) that enables a commentary agent based on LLMs to deliver detailed commentaries on cooperation and strategic interactions with opponent agents in Guandan, a card game of incomplete information, without the need for specialized training in Chinese text environments. This task is broken down into several core components, including a State Commentary Guider, ToM Strategy Analyzer, and Style Retriever. The inputs received by the LLM include game rules, observation rules (i.e., prompts that guide the LLM to convert low-level game state information into readable text), and historical game context. We provide a detailed description process demonstrating how to guide the LLM to utilize its knowledge, reasoning capabilities, and ToM to perform these module functions and effectively navigate the complexities inherent in incomplete information games.

\vspace{-0.5em}
\subsection{State Commentary Guider}
\vspace{-0.5em}
In the commentary of the Guandan game, we propose a State Commentary Guide module aimed at assisting the language model in transforming the game state into readable commentary text. This module integrates game rules, observation transformation rules, and historical transformation rules, guiding the model to convert the current low-level state and historical information into descriptive text to support the commentary process. Please refer to the appendix~\ref{app:a} for detailed rules.
We design structured prompts to help the LLM understand the game rules and current state information, including card types, scoring rules, and win/loss conditions. An example of the game rule template is as follows: The \textbf{game rule} template is shown as follows: 

\begin{itemize}[leftmargin=10pt]
    \item\textbf{ General Rules:} A brief game introduction, team position rules, all single cards, and cards ranking.
    \item \textbf{Card Type that cannot beat other types:} \{Description of Card Type 1, Example of Card Type 1\}, \{Description of Card Type 2, Example of Card Type 2\}, ...;
    \item \textbf{Card Type that can beat other types:} \{Description of Card Type 1, Example of Card Type 1\}, \{Description of Card Type 2, Example of Card Type 2\}, ...;
    \item \textbf{Single Win/Loss Rule:} The scoring rules for four players in a single game, with different combinations of cards being played out in different order.
    \item \textbf{Whole Win/Loss Rule:} The overall win/loss conditions.
\end{itemize}


We also design templates for observation rules and historical transformation rules, which include input interpretations and transformation prompts. Through these templates, the model can convert low-level game states and history into vivid commentary text, making it easier for the audience to understand the game's progression and players' strategies. The template is shown as follows:

\begin{itemize}[leftmargin=10pt]
    \item \textbf{Input Explanation:} The input types, like dictionaries and lists, are clearly specified. A description of every component in the input is also provided.
    \item \textbf{Conversion Tips:} Additional instructions guide converting the low-level game state representations into natural language descriptions.
\end{itemize}

We can effectively convert low-level game states and history records into human-readable text, denoted as $\textit{Obs}_r$ and $\textit{His}_r$, respectively, by employing the game rule, observation conversion rule, and history conversion rule. This process enhances the clarity and comprehensibility of the game commentary, providing more vivid and understandable content for the audience. The conditional distribution for each element $\textit{Obs}_r[j]$ within the generated text can be modeled using the prompts $\textit{Prompt}_{obs}$ as: 
\vspace{-0.3em}
\begin{equation}
\mathbb{P}(Obs_r) \sim \prod_{j=1}^{N} LM_{\theta}\big(Obs_r[j] \mid Prompt_{obs}, Rule, Rule_{obs}, Obs_r[1, \ldots, j-1]\big)
\end{equation}

$\textit{LM}_{\theta}$ represents the language model parameterized by $\theta$, and $\textit{N}$ is the length of the generated text $\textit{Obs}_r$. This State Commentary Guide module provides crucial support for the model to commentate in games of incomplete information.
Here we want to provide a theoretical guarantee for your mechanism design.

\begin{theorem}[Existence of Compliant Commentary Generation]
\label{thm:compliant_commentary}
Let $S$ be a finite game state space, and let $Rule$, $Rule_{obs}$, and the history conversion rule form a finite set of conversion rules.
Then, there exists a language model $\mathrm{LM}_{\theta}$ such that, with probability at least $1 - \varepsilon$, $\text{LM}_{\theta^*}$ produces compliant narrations consistent with the rules for any state $s \in S$ and its corresponding histories. Formally,
\begin{equation}
\mathbb{P}\Bigl(\text{LM}_{\theta^*}\bigl(s, \textit{His}_r\bigr) \text{ is compliant}\Bigr)\ge 1 - \varepsilon.
\end{equation}

\end{theorem}
In particular, this guarantees that the State Commentary Guide module can effectively support commentary in games of incomplete information by generating human-readable observations that respect the specified rules. The detailed background and underlying proof are provided in the~\cref{sec:thm}.


\vspace{-0.5em}
\subsection{TOM-Based Strategy Analyzer}
\vspace{-0.5em}
In the commentary of the card game, Guandan, the psychological tactics and strategic interactions within the game are difficult to accurately assess from a single player's perspective. Therefore, we have adopted the ToM~\cite{ToM,DBLP:journals/corr/abs-2302-02083,Suspicion-Agent} approach to enhance the depth and strategy of the commentary. The commentary model is based on processing incomplete information of the game and inferring the psychological states of players, thereby providing accurate analysis and predictions in a complex multi-player interaction environment.

\noindent\textbf{First-order ToM:} 
We utilize first-order ToM for basic strategy analysis. By analyzing players' past actions and the current situation, we infer the possible hand types each player may hold and analyze their potential strategies. This information is utilized to construct commentary content, explaining players' actions and potential counter-strategies by opponents. For example, if a player consistently chooses to play certain cards, we can infer that they might hold strong cards and possibly intend to suppress their opponents.

\noindent\textbf{Second-order ToM:} 
We introduce second-order ToM for a deeper level of strategy analysis. At this stage, we not only consider players' strategies and actions but also predict opponents' cognition and reactions to these strategies. Through this approach, we can interpret the game progression and players' strategic tendencies more comprehensively. For instance, if a player adopts a relatively adventurous strategy in a certain situation, we may speculate that they believe their opponents would not anticipate this move, deliberately selecting this strategy.

To ensure the logical coherence and fluency of the commentary content, after stepwise reasoning on strategies, we introduce a Commentary Coordinator to review the output of the game commentary. The Commentary Coordinator is responsible for integrating various commentary parts, ensuring the content is both accurate and easy to understand, and seamlessly blending with the game progression. Through this method, we can provide audiences with deeper game commentary, allowing them to better understand the strategic competition and psychological tactics among players.

\vspace{-0.5em}
\subsection{Style Retrieval and Extraction}
\vspace{-0.5em}
We introduce a Style Retrieval and Extraction module specifically designed for the Guandan game to enhance the accuracy and relevance of information in the game commentary system. The module is divided into two main stages: data retrieval and information filtering.

\noindent\textbf{Data Retrieval:} 
In the data retrieval phase of Guandan game commentary, we adopt a tree structure retrieval method to efficiently extract information most relevant to user queries from a corpus specifically designed for the Guandan game. In this process, each document is broken down into individual document nodes, each containing the content of the original document and a unique identifier, to facilitate subsequent vector indexing. The content of these document nodes is then converted into vector form and indexed using a vector space model.

During the query execution phase, the system first converts the user's query request into vector form to ensure that the query can be effectively compared with the vectorized document nodes. Then, the system searches the constructed vector index, returning only document nodes whose similarity exceeds a preset threshold. Our threshold is set to 2, meaning that the system filters out nodes with a similarity greater than this value, ensuring that only the most relevant information is extracted.

\noindent\textbf{Information Filtering:} 
In the information filtering stage, the model meticulously filters the retrieved data. The system examines the relevance of each data item, retaining only those that meet high standards for subsequent processing. The content outputted by the Style Retrieval and Extraction module will provide curated, relevant data to the Commentary Coordinator to support the analysis and explanation of game situations and player strategies during the commentary process.



\vspace{-0.5em}
\section{Experiments and Result Analysis}
\vspace{-0.5em}
\begin{figure*}[!ht]
  \centering
  \includegraphics[width=1\textwidth]{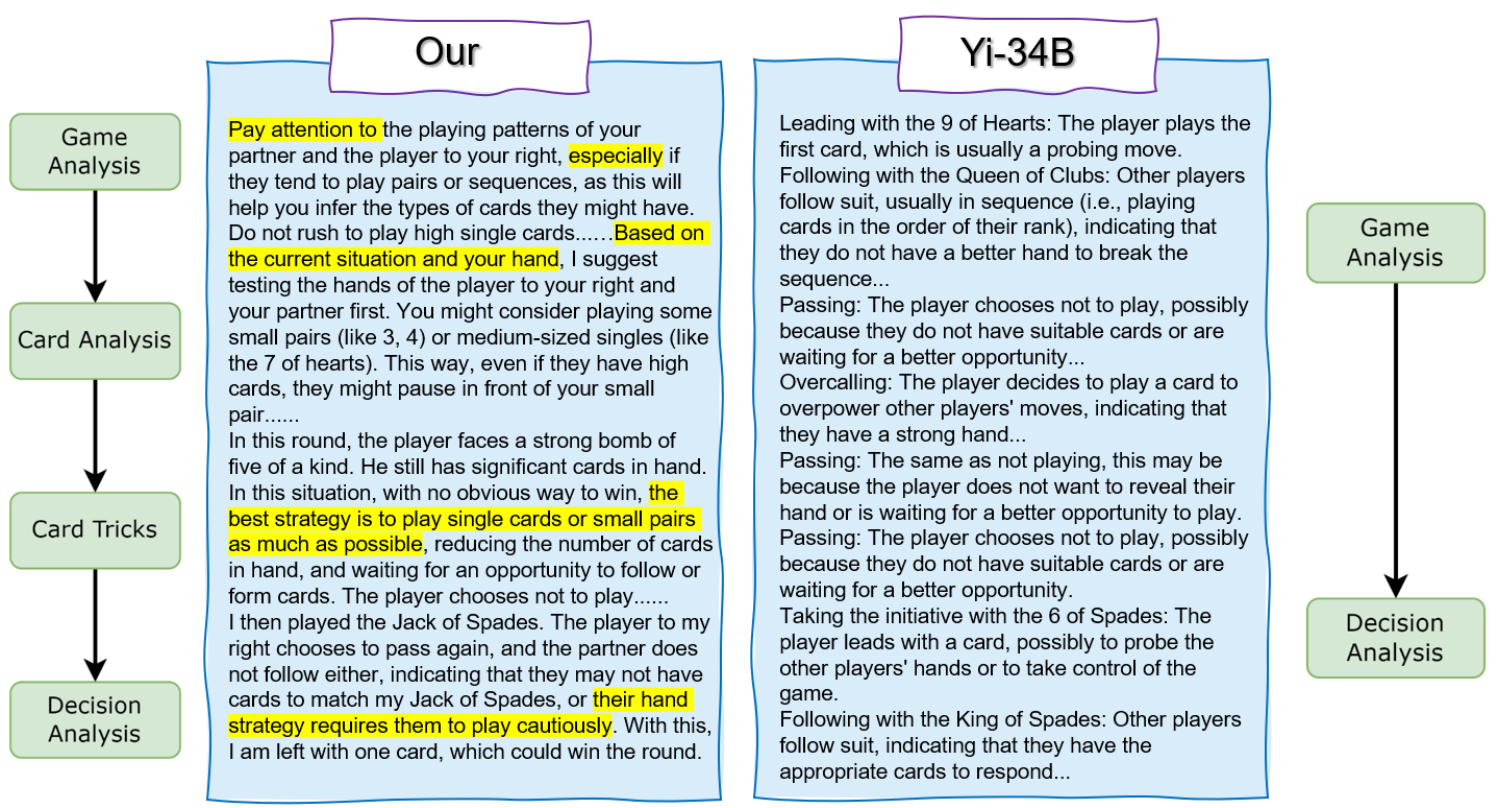}
     \captionof{figure}{
Simulate multi-round outputs for different methods of game commentary.}
  \label{fig:case}
\end{figure*}
\vspace{-0.5em}
\subsection{Implementation details}
\vspace{-0.5em}
To study the performance of an LLM-based agent in Guandan commentary under conditions of incomplete information, lack of communication, and dynamic collaboration, we choose Guandan as the experimental environment. In our tests, we use \textbf{Danzero+} ~\cite{zhao2023danzero+} as the game agent. Danzero+ employs a distributed framework to train the reinforcement learning model using the deep Monte Carlo method. They demonstrate their capability to perform at a human level. All experiments are conducted in a Chinese environment.We conducted tests on both open-source and close-source language models, including OpenAI’s commercial model GPT-4 and GPT-3.5~\cite{achiam2023gpt}, and Chinese-language open-source models like Qwen-32B~\cite{bai2023qwen}, Yi-34B~\cite{ai2024yi}, ChatGLM-4~\cite{zeng2022glm}.

In the data processing stage, we first remove non-text elements and stopwords from the texts to reduce noise. Then, we conducted text normalization, including unifying the case of the text to eliminate differences caused by case inconsistency. We also applied part-of-speech tagging and tokenization techniques. Finally, we use the Porter Stemmer algorithm to simplify word inflections.~\cite{porter1980algorithm}

\vspace{-0.5em}
\subsection{Dataset}
\vspace{-0.5em}
Our dataset includes professional commentary texts and generated commentary texts, detailed as follows:

\noindent\textbf{Professional commentary:} 
 Collected from the transcripts of Guandan match commentary videos, provided by experienced commentators with unique personal styles. These texts display high-level strategic insights and rich contextual descriptions, reflecting the personalized expression and professional skills of the commentators.
 
\noindent\textbf{Generated commentary:}Produced by the commentary model based on actual match data, simulating the style and content of professional commentary.
\subsection{Metrics}
\vspace{-0.5em}
     To comprehensively evaluate the quality and applicability of the commentary texts, we use the following metrics:
     
    \noindent\textbf{Cosine Similarity:} 
    We employ the TF-IDF vectorization method~\cite{aizawa2003information} to convert processed texts into vector form and calculate the cosine similarity between professional and test texts. This measures the semantic closeness of professional and generated texts, reflecting the model's capability to capture semantics.
    
    \noindent\textbf{Sentiment Analysis:} 
    Through sentiment analysis~\cite{medhat2014sentiment} tools, we assign sentiment polarity scores to the texts to compare the emotional expression differences between the two types of texts.
    
    \noindent\textbf{Lexical Diversity:} 
    We use the Type-Token Ratio (TTR)~\cite{richards1987type} to assess the lexical diversity of the texts. This metric measures the proportion of different words in the text relative to the total number of words, reflecting the richness and diversity of the language.

    \noindent\textbf{SNOWNLP:} 
    The SNOWNLP~\cite{SnowNLP} score is a sentiment score ranging from 0 (most negative) to 1 (most positive), computed using a Chinese text sentiment analysis library based on Naive Bayes.
    
    \noindent\textbf{Human Evaluation:} 
    We also conduct human evaluations, including match consistency and fluency. These evaluations are completed by human reviewers to verify whether the commentary texts accurately reflect the actual card game situations and assess the readability and naturalness of the texts. Reviewers need to have a certain knowledge and experience of Guandan to accurately judge the accuracy of the text descriptions.

\begin{enumerate}[leftmargin=10pt]
\item \textbf{Match Consistency}
\begin{itemize}[leftmargin=10pt]
    \item \textbf{Key Event Identification(KEI):} Assess whether the commentary texts capture key events in the match, such as major turnarounds or critical decision points~\cite{renella2023towards}.
   \item \textbf{Detail Accuracy:} Check if the text descriptions of card types, scores, and player strategies are precise and correct.
\end{itemize}

\item \textbf{Fluency}
\begin{itemize}[leftmargin=10pt]
    \item 
    \textbf{Naturalness:}Assess if the text language is smooth and free of grammatical errors or unnatural expressions.
    \item \textbf{Information Organization:} Evaluate if the text's information is properly organized and can be understood in a logical order throughout the development of the match.
   \item \textbf{Logical Coherence:}Check if the narrative of events in the text is coherent and free from logical jumps or contradictions.
\end{itemize}
\end{enumerate}
\vspace{-0.5em}
\subsection{Results}
\vspace{-0.5em}

\begin{table*}  
\centering
\caption{Evaluation Results of Commentary for Open Source and Closed Source Models}
\renewcommand{\arraystretch}{1.2} 
\resizebox{\textwidth}{!}{
\begin{tabular}{lccccccc}
\hline  \hline 
\multicolumn{1}{c}{} & \multicolumn{4}{c}{\textbf{Sentiment Analysis}} & \multirow{2}{*}{\textbf{Cosine Similarity}} & \multirow{2}{*}{\textbf{Lexical Diversity}} &  \multirow{2}{*}{\textbf{SNOWNLP}} \\
\cline{2-5}
 & \textbf{neg}& \textbf{neu}&\textbf{pos}& \textbf{compound}&  &  & \\
\hline
GPT-3.5 &  0.0 & 1.0 & 0.0 & 0.0 & 0.0032 & 0.09 & 0.0\\
GPT-4 &  0.0 & 1.0 & 0.0 & 0.0 & 0.0380 & 1.0 & 0.0\\
Yi-34B &  0.0 & 1.0 & 0.0 & 0.0 & 0.0250 & 0.55 & 0.99 \\
GLM-4 &  0.0 & 1.0 & 0.0 & 0.0 & 0.0050 & 0.86 & 0.98\\
Our & 0.0 & 1.0 & 0.0 & 0.0 & 0.7955 & 0.95 & 0.0 \\
\hline 
Original & 0.0 & \textbf{1.0} & 0.0 & 0.0 & - & \textbf{1.0} & 0.0 \\
\hline \hline 
\end{tabular}}
\label{tab:Result}    
\end{table*}

\begin{table*}  
\centering
\caption{Human Evaluation of different benchmarks and our model}
\renewcommand{\arraystretch}{1.2} 
\resizebox{\textwidth}{!}{
\begin{tabular}{lcccccc}
\hline \hline 
\multicolumn{1}{c}{} & \multicolumn{2}{c}{\textbf{Match Consistency}} & \multicolumn{1}{c}{} & \multicolumn{3}{c}{\textbf{Fluency}} \\
\cline{2-3} \cline{5-7}
 & \textbf{KEI}&\textbf{Detail Accuracy}& & \textbf{Naturalness}& \textbf{Information Organization} & \textbf{Logical Coherence} \\
\hline
GPT-3.5   &0.37  &0.95 & &0.86 &0.69  &3.75    \\
GPT-4   &0.46  &0.91 & &0.90&0.80  & 4.46   \\
Yi-34B   &0.23  &0.87 & &0.82  &0.62  & 3.52 \\
GLM-4   & 0.32 & 0.83& & 0.84 &0.58  & 2.72  \\
Our   & \textbf{0.81} &  \textbf{0.97}& &  \textbf{0.95} & \textbf{0.89} & \textbf{4.34}  \\
\hline  \hline 

\end{tabular}}
\label{tab:human}    
\end{table*}

In this experiment, we conduct a comparative analysis of data from Table~\ref{tab:Result}, revealing that utilizing the Retrieval-Augmented Generation (RAG) framework for commentary provides substantial benefits. By integrating natural and personable colloquialisms and key techniques such as card memory, the retrieval mechanism in RAG delivers a more authentic and professionally styled commentary. Even with its larger parameter count, GPT-4 without retrieval still falls short in accuracy compared to RAG-enabled models. The retrieval support of RAG helps commentators offer more precise analyses and predictions, giving viewers deeper insights into match progress and creating a stronger emotional connection to the original text.

When applied to card game commentary (e.g., Guandan), RAG incorporates additional steps for Chain-of-Thought (CoT) retrieval. Equipping large models with such retrieval features allows them to more faithfully emulate the linguistic norms of expert commentators, outperforming large models without retrieval or those simply fine-tuned on commentary data.

All evaluated models exhibit neutrality in sentiment analysis. GPT-3.5 shows relatively low lexical diversity (0.09), whereas Qwen equipped with RAG and GPT-4 both yield higher diversity in generated text. The near-perfect SnowNLP score (0.99) of the Yi-34B model reflects neither a clear advantage nor a reliable mimicry ability. In addition, all models struggle with cosine similarity against the original text, likely due to the improvisational nature of live commentary. Among them, Qwen with RAG registers a slight edge in aligning with the source content.

\begin{table*}  
\centering
\caption{Ablation study on the effect of Tree-Based Retrieval Method}
\renewcommand{\arraystretch}{1.3} 
\setlength{\tabcolsep}{4pt} 
\resizebox{\textwidth}{!}{ 
\begin{tabular}{lccccccc}
\hline \hline
 & \multicolumn{4}{c}{\textbf{Sentiment Analysis}} & \textbf{Cosine} & \textbf{Lexical} & \textbf{SNOWNLP} \\
\cmidrule(lr){2-5} \cmidrule(lr){6-6} \cmidrule(lr){7-7} \cmidrule(lr){8-8}
 & \textbf{neg} & \textbf{neu} & \textbf{pos} & \textbf{compound} & \textbf{Similarity} & \textbf{Diversity} & \\
\hline
Our(w/o RAG)(Vanilla) & 0.0 & 0.0 & 1.0 & 0.0 & 0.0 & 0.87 & 1.0 \\
Our(w/o RAG)(1st-ToM) & 0.0 & 0.0 & 1.0 & 0.0 & 0.0126 & 1.0 & 1.0 \\
Our(w/o RAG)(2nd-ToM) & 0.0 & 1.0 & 0.0 & 0.0 & 0.0380 & 1.0 & 1.0 \\
Our(w RAG)(1st-ToM) & 0.0 & 1.0 & 0.0 & 0.0 & 0.7519 & 0.92 & 0.0 \\
Our(w RAG)(2nd-ToM) & 0.0 & 1.0 & 0.0 & 0.0 & 0.7955 & 0.95 & 0.0 \\
\hline 
Original & 0.0 & 1.0 & 0.0 & 0.0 & - & 1.0 & 0.0 \\
\hline \hline
\end{tabular}}
\label{tab:ablation}    
\end{table*}

\noindent\textbf{Human Evaluation Analysis.}
We recruit 20 human annotators with Guandan experience for scoring. As shown in Table~\ref{tab:human}, in terms of match consistency, our model significantly outperforms other models, particularly excelling in Key Event Identification (KEI)~\cite{lin2002manual}. This demonstrates that our model accurately captures crucial moments in the game, effectively reflecting the game's turning points and climactic sections. In detail accuracy, our model also performs exceptionally well, accurately describing game elements such as card types, scores, and strategic actions. Regarding fluency, our framework scores high in naturalness (0.95), information organization (0.89), and logical coherence (4.34), highlighting the model's ability to generate commentary text that is grammatically correct, logically structured, and contextually coherent.

In contrast, other models like GPT-4, despite performing well in general language generation tasks, often fail to show the same acuity in specialized game commentary scenarios. These models are often not sensitive enough to key events, sometimes missing significant turning points in the game or failing to fully utilize game-specific terminology and expressions. Additionally, in terms of fluency, although they can generate structurally sound sentences, they sometimes lack the ability to present information in a logical and captivating manner.

\noindent\textbf{Case Study.} As shown in the figure~\ref{fig:case}, our method not only focuses on the types of cards played but also delves into detailed analysis of the playing patterns and potential strategies of the players and their opponents. It provides recommendations based on the current and predicted game state, advising when to play high or low cards. This includes discussions on when to play single cards or pairs to maximize gameplay advantage, helping players not only understand the game rules but also master advanced strategies to enhance their gameplay level. In contrast, the Yi-34B model's commentary seems to focus more on the sequence of cards played without deeply exploring the reasoning or strategic implications behind these choices. It lists actions such as leading with certain cards or choosing to pass, which, while informative, do not delve into the strategic significance of these decisions.See Appendix~\ref{app:b}  for more output examples.
\vspace{-0.5em}
\subsection{Ablation Studies}
\vspace{-0.5em}

\noindent\textbf{Ablation Studies on RAG.}
Table~\ref{tab:ablation} shows that removing RAG (e.g., “Our(w/o RAG)(Vanilla)”) results in a cosine similarity of 0.0 with the original text, despite producing the maximum SNOWNLP score of 1.0 and a reasonably high lexical diversity (0.87 or 1.0). This outcome implies that, without retrieval, the generated text deviates significantly from the source content, even though it may appear stylistically diverse. In contrast, when RAG is introduced (“Our(w RAG)(1st-ToM)” or “Our(w RAG)(2nd-ToM)”), the cosine similarity rises markedly (up to 0.7955), indicating stronger alignment with the original text. However, this improvement is accompanied by a slight decrease in both lexical diversity and SNOWNLP scores, suggesting that retrieval imposes some constraints on free-form generation. Overall, these results underscore the importance of RAG for achieving higher semantic fidelity in commentary.

\noindent\textbf{Ablation Result Analysis.}  
The RAG model with the retrieval component displays significant disparities from the model lacking it across various dimensions. In terms of lexical diversity, the model incorporating retrieval exhibits somewhat constrained diversity, suggesting a potential limitation imposed by the retrieval component. Sentiment analysis using the SnowNLP tool reveals that the model without retrieval yields more pronounced sentiment results, diverging notably from the original text. This deviation may arise from the model's greater freedom in generating emotional expressions, albeit resulting in a less faithful imitation of the source text. Conversely, regarding text semantic similarity, the model integrating retrieval showcases a distinct advantage, effectively highlighting the crucial role of the retrieval component in maintaining semantic coherence and bolstering text relevance.

In addition to retrieval, second-order ToM demonstrates a clear advantage over first-order ToM in semantic alignment. Without RAG, second-order ToM improves cosine similarity from 0.0126 to 0.0380, and, with RAG, from 0.7519 to 0.7955. These gains indicate that higher-order ToM facilitates a deeper understanding of strategic interactions and player intentions, enhancing the commentary’s accuracy. In Guandan, for instance, second-order ToM captures opponents’ potential counteractions and anticipations, resulting in more precise narration of pivotal turning points. As such, second-order ToM is instrumental in producing in-depth and contextually grounded game commentary.

\vspace{-0.5em}
\section{Conclusion}
\vspace{-0.5em}

In this study, we introduce a novel commentary framework that combines reinforcement learning and large language models to generate detailed, context-relevant commentary for the complex, incomplete information card game Guandan. Our modular approach, consisting of a State Commentary Guider, ToM Strategy Analyzer, and Style Retrieval module, enables the LLM to deliver insightful game commentary without the need for specialized training. Experimental results demonstrate that our commentary framework significantly enhances the performance of open-source LLMs, outperforming GPT-4 on multiple evaluation metrics. In the future, we aim to extend our commentary framework to other complex games and explore the integration of additional modalities, such as audio and video data, to further enrich the commentary generation process.

\bibliography{custom}
\bibliographystyle{colm2025_conference}
\clearpage

\appendix

\section{proof of Theorem~\ref{thm:compliant_commentary}}\label{sec:thm}

The theory of mechanism design is crucial for the architecture of large language models (LLMs) framework design, as it provides a principled framework to align model outputs with desired objectives through incentive-compatible structures. Below is a detailed proof of Theorem~\ref{thm:compliant_commentary} based on mathematical mechanism design, demonstrating that the Compliant Commentary Generation is feasible.

First, we want to write a road map for this proof.
\begin{itemize}[leftmargin=10pt]
    \item We encode each state-rule pair \( (s_t, R) \) as a vector \( v_t \). Then we prove that the covered space \( V \subset \mathbb{R}^d \) is compact (since both \( S \) and \( R \) are finite).

    \item We we want to prove that by the universal approximation theorem~\cite{hornik1991approximation}, there exists a neural network \(\text{LM}_\theta \) such that:
    \begin{equation}
\sup_{v \in V} \left\| \mathrm{LM}_\theta(v) - \mathbb{P}(c \mid v) \right\| < \varepsilon
\end{equation}
where \( c \) denotes a Compliant Commentary Generation.
    \item The generation process forms a Markov chain whose stationary distribution \( \pi(c) \) satisfies:
    \begin{equation}
        \pi(c) = \sum_{v} \mathbb{P}(c \mid v)  \mathbb{P}(v)
    \end{equation}
    \item After that, the convergence theorem is guaranteed by the Perron-Frobenius theorem~\cite{horn1985matrix}, 
    
    \begin{equation}
\mathbb{P}\Bigl(\text{LM}_{\theta^*}\bigl(s, \textit{His}_r\bigr) \text{ is compliant}\Bigr)\ge 1 - \varepsilon.
\end{equation}
    which is Theorem~\ref{thm:compliant_commentary}.  
\end{itemize}

Let's start the proof.

\begin{lemma}\label{lemma:1}
Let \( S \) be a finite state space and \( R \) a finite rule set. Then each state-rule pair \( (s_t, R) \) can be encoded as a vector \( v_t \in \mathbb{R}^d \), such that the space of all encoded vectors \( V \subset \mathbb{R}^d \) is compact.
\end{lemma}
\begin{proof}

We wish to show that the set  
\begin{equation}
V = \{v_t \in \mathbb{R}^d \mid v_t \text{ encodes the state-rule pair } (s_t,R), s_t\in S, R \in \mathcal{R}\}
\end{equation}
is compact, since both $S$ and $\mathcal{R}$ are finite. The Cartesian product $S \times \mathcal{R}$ is finite. The encoding function
\begin{equation}
f: S \times \mathcal{R} \to \mathbb{R}^d,\quad (s_t, R) \mapsto v_t
\end{equation}
assigns a vector $v_t$ in $\mathbb{R}^d$ to each state-rule pair. Therefore, the set $V = f(S \times \mathcal{R})$
must be finite because it is the image of a finite set under the function $f$.

Any finite subset of $\mathbb{R}^d$ is bounded. In particular, there exists some $M > 0$ such that for every $v_t \in V$, the norm satisfies
$\|v_t\| \le M.$ A finite set has no limit points other than its own elements, and every sequence in a finite set eventually becomes constant (or repeats elements). Thus, all limit points of $V$ are contained in $V$ itself, implying that $V$ is closed.

Then, in $\mathbb{R}^d$, the Heine-Borel theorem~\cite{rudin1976principles} tells us that a subset is compact if and only if it is closed and bounded. Since we have shown that $V$ is both closed and bounded, it follows that $V$ is compact.
\end{proof}
\begin{lemma}
\label{lemma:universal_approx}
Let \( V \subset \mathbb{R}^d \) be a compact set, and let \( \mathbb{P}(c \mid v) \) denote the conditional distribution over compliant commentary \( c \) given input vector \( v \in V \). Then, by the universal approximation theorem~\cite{hornik1991approximation}, there exists a neural network \( \mathrm{LM}_\theta \) such that:
\begin{equation}
\sup_{v \in V} \left\| \mathrm{LM}_\theta(v) - \mathbb{P}(c \mid v) \right\| < \varepsilon,
\end{equation}
for any \( \varepsilon > 0 \), where \( \mathrm{LM}_\theta(v) \) approximates the target distribution \( \mathbb{P}(c \mid v) \).
\end{lemma}
\begin{proof}

Assume that the target function $f(v) = \mathbb{P}(c \mid v),$
which returns the probability (or an appropriate representation) of a Compliant Commentary Generation $ c $ based on the encoding $ v $, is continuous on the compact set $ V $.

The universal approximation theorem (see \cite{hornik1991approximation}) asserts that for any continuous function defined on a compact subset of $\mathbb{R}^d$, and for any $\varepsilon > 0$, there exists a feedforward neural network $ \mathrm{LM}_\theta $ with appropriate architecture and parameters $\theta$ such that
\begin{equation}
\sup_{v \in V} \left\| \mathrm{LM}_\theta(v) - f(v) \right\| < \varepsilon.
\end{equation}
Substituting $ f(v) = \mathbb{P}(c \mid v) $ yields
\begin{equation}
\sup_{v \in V} \left\| \mathrm{LM}_\theta(v) - \mathbb{P}(c \mid v) \right\| < \varepsilon.
\end{equation}

\end{proof}

\begin{lemma}[Theorem~\ref{thm:compliant_commentary}]
\label{lemma:markov_convergence}
Let \( c \) denote a compliant commentary generation. Suppose the generation process induced by the language model \( \mathrm{LM}_\theta \) forms a Markov chain over the commentary space. Then the stationary distribution \( \pi(c) \) of this process is given by:
\begin{equation}
    \pi(c) = \sum_{v} \mathbb{P}(c \mid v) \, \mathbb{P}(v),
\end{equation}
where \( v \) denotes the encoded state-rule representations. Furthermore, by the Perron--Frobenius theorem~\cite{horn1985matrix}, the Markov chain converges to \( \pi(c) \), and the learned model \( \mathrm{LM}_{\theta^*} \) satisfies:
\begin{equation}
    \mathbb{P}\Bigl(\mathrm{LM}_{\theta^*}\bigl(s, \textit{His}_r\bigr) \text{ is compliant}\Bigr) \ge 1 - \varepsilon.
\end{equation}
\end{lemma}

\begin{proof}

Assume that the commentary generation process evolves over discrete time steps $ t $ such that the generated commentary $ c_t $ only depends on the previous commentary $ c_{t-1} $. Thus, the process forms a Markov chain with the state space being the set of all compliant commentaries. Let the transition probabilities be given by
\begin{equation}
\mathbb{P}(c' \mid c) = \mathbb{P}(c_{t+1} = c' \mid c_t = c),
\end{equation}

there exists a unique stationary distribution $\pi(c)$ that satisfies
\begin{equation}
\pi(c') = \sum_{c} \pi(c) \, \mathbb{P}(c' \mid c).
\end{equation}
This stationary distribution represents the long-term behavior of the chain. Then, the Perron-Frobenius theorem applied to the transition probability matrix $ P $ (a non-negative matrix) guarantees that: (1)The spectral radius of $ P $ is 1. (2) There is a unique positive eigenvector (up to scaling) corresponding to the eigenvalue 1.

Assume that the training of the neural network $\mathrm{LM}_{\theta^*}$ is designed so that the target (or true) process overwhelmingly produces compliant commentaries. In other words, the intended stationary distribution $\pi(c)$ for a compliant commentary $c$ satisfies
\begin{equation}
\pi(c) \geq 1-\varepsilon.
\end{equation}
This is a consequence of the training objective and approximation guarantees (e.g., from the universal approximation theorem), ensuring that the probability of compliance asymptotically is at least $1-\varepsilon$.

Since the generation process is implemented via the neural network $\mathrm{LM}_{\theta^*}$, we have by the convergence property of the Markov chain:
\begin{equation}
\lim_{t \to \infty} \mathbb{P}\Bigl( \mathrm{LM}_{\theta^*}(s,\textit{His}_r) \text{ produces } c \Bigr) = \pi(c).
\end{equation}
Thus, when the system has run for a sufficiently long time, the probability that $\mathrm{LM}_{\theta^*}(s,\textit{His}_r)$ produces a compliant commentary is at least $1-\varepsilon$:
\begin{equation}
\mathbb{P}\Bigl( \mathrm{LM}_{\theta^*}(s, \textit{His}_r) \text{ is compliant} \Bigr) \ge 1-\varepsilon.
\end{equation}

\end{proof}

\section{Guandan Rules} \label{app:a}

\subsection{Basic Rules}\label{app:a1}
\begin{mybox1}
\textbf{English}
\par
\medskip
\textit{1. Players are divided into 2 teams, with teammates sitting opposite each other. The game uses two decks of playing cards.}

\textit{2. A complete game of Guandan consists of several rounds, each with its own trump card. The trump card is the highest card below the jokers. For example, if the trump card is 2, the order of cards from lowest to highest is 3, 4, 5, 6, 7, 8, 9, 10, J, Q, K, A, 2. If the trump card is 5, the order is 2, 3, 4, 6, 7, 8, 9, 10, J, Q, K, A, 5.}

\textit{3. Winning a round: The first team whose member finishes their cards wins the round, but the round does not end until the order of finishing is determined, meaning three players must finish their cards.}

\textit{4. Trump card calculation rule 1: The initial trump card value is 2, which then increases based on the win/loss outcomes of the teams.}

\textit{5.Trump card calculation rule 2: Each team has its own trump card, starting from 2.}

\textit{6. Trump card calculation rule 3, upgrades from 1-3 levels. If you are the first player to finish your hand in a round, your team wins the round, and your team’s trump card is upgraded (increased in value). The specific upgrade depends on your teammate's finishing position. Best victory: if your teammate finishes second, upgrade by 3 levels (e.g., from 2 to 5); third finisher upgrades by 2 levels; fourth finisher upgrades by 1 level.}

\textit{7. Winning the whole game of Guandan: A team that upgrades their trump card to A and achieves the best victory in that round wins the entire game of Guandan. The game only concludes completely then. If the team has a trump card of A, that round becomes a critical challenge, and the team has three chances (which can be non-consecutive rounds) to achieve the best victory. If they fail in these three rounds, the trump card resets to 2.}

\textit{8. Lead team rotation: The lead team for a round is determined by whose trump card is in play. If that team wins the round (not necessarily the best victory), they retain the lead team status, and their trump card is used in the next round. Otherwise, if the opposing team wins, they become the lead team, and their trump card is used in the next round. Regardless of which team wins the round, the winning team's trump card is upgraded according to rule 3 and applied in the next round.}

\textit{9. Tribute (card handover): \textcolor{blue}{After the two players who finished their cards first, regardless of whether they are teammates or not, the two players who finished their cards later must surrender the highest card.} If a player who needs to surrender cards has two Big Jokers, they can resist surrendering. Situation one: in the previous round, the top two players from the same team finished first, and each player from the opposing team contributes their highest card. Situation two: for other situations, follow the instructions in blue.}

\textit{10. Tribute rule supplement 1: Dual jokers can refuse tribute (jokers need not be held by the same person in scenario one).}

\textit{11. Tribute rule supplement 2: The first finisher can choose one of two cards offered in tribute. Whichever card they choose, that player starts the next round.}

\textit{12. Tribute rule supplement 3: The player receiving tribute must return a card to the tribute giver, which can be semi-freely chosen (only a card valued 10 or lower; if all cards are above 10, the smallest must be returned).}

\textit{13. Tribute rule supplement 4: The tribute must include the highest card, excluding jokers and the trump card, unless the trump card is a heart.}

\textit{14. Wild card: The heart suit of the trump card (two cards) can act as any card except the jokers. For example, if the trump card is 6, then the heart 6 can be used as any other card.}

\medskip

\end{mybox1}

\clearpage

\begin{mybox1}
\textbf{Chinese}
\begin{CJK}{UTF8}{gkai}
\par
\medskip

1. 四个人分 2 队，同一队内的两人对坐，两幅扑克。

2. 一整场掼蛋会有多局游戏，每一局都有自己的主牌。主牌是大小王之下最大的牌，比如主牌是 2，那牌面从小到大就是 3 4 5 6 7 8 9 10 J Q K A 2，如果主牌是 5，那牌面从小到大是 2 3 4 6 7 8 9 10 J Q K A 5。

3. 一局的输赢规则：第一个先走完牌的队员所属的那一队赢，但是此局仍未结束，要等到确定了出完牌的排名次序才结束，也即要有三个人出完牌才结束。

4. 主牌计算的规则 1。主牌初始值是 2，随后根据双方队伍输赢情况增加主牌的数值。

5. 主牌计算的规则 2。两支队伍有各自的主牌（需要保存），但都是从 2 开始。

6. 主牌计算的规则 3，升级数 1-3。如果您是当局第一个出完手牌的玩家，那么您的队伍获得当局的胜利，您的队伍的主牌将会获得升级（数值变大）。当然，主牌的升级数具体看队友出完牌的排位。最佳胜利：队友是第 2 个走完的，升 3 级（如果此前主牌是 2，那现在变成 5）；第 3 个走完升 2 级；第四个走完升 1 级。

7. 一整场掼蛋的输赢规则：当一支队伍把主牌升级到 A，并且在 A 的那一局仍然取得最佳胜利，那么那支队伍获得整场掼蛋的胜利，此时的掼蛋游戏才算完整结束。当然，主牌为 A的那局对于对应的队伍而言是渡劫局，该队伍总共有三次机会（可以是不连续的三局），如果那三局都没有取得最佳胜利，那么主牌会重新回到 2 开始。

8. 主队轮换：一局游戏的主牌是哪支队伍的，那支队伍就是这一局游戏的主队。如果那支队伍在这局游戏取得胜利（不必是最佳的），那么他们能够保留主队身份，下一局游戏仍然使用他们的主牌。否则，一旦敌对队伍取得胜利，那么主队就变成了敌方队伍，下一局的主牌就是使用敌方队伍的。无论取得某一局胜利的是不是主队，局胜队伍的主牌都要根据规则3 进行升级，并且在下一局应用。

9. 进贡（缴牌）。\textcolor{blue}{后出完牌的两位向先出完牌的两位缴牌（无论是不是队友），必须是最大的那张。}如果需要缴牌的玩家拥有两张大王则可以抗缴。情况一：上一局同一队 2 人是前 2 名走完，敌方队伍双方各自进贡最大的 1 张。情况二：其他情况，按照蓝字处理。

10. 缴牌规则补充 1。双大王可抗贡（情况一大王可不在同一人手中）。

11. 缴牌规则补充 2。第 1 个出完的玩家可以在进贡的牌二选一。拿谁的，下一局谁先出牌。

12. 缴牌规则补充 3。被进贡的玩家必须还一张手牌给进贡者，这张手牌可以半自由选择。
（只能还不超过 10 的牌，如果全手牌都大于 10，还张最小的）

13. 缴牌规则补充 4。进贡要进最大的牌，没有大小王，有主牌，就进贡主牌，但是红桃主牌除外。

14. 万能牌：主牌的红心花色（两张），可以充当任意除大小王之外的牌。比如主牌是 6，那么红心 6 可以当做任意一张牌使用，除了大小王。

\medskip
\end{CJK}
\end{mybox1}

\clearpage

\subsection{Additional Rules}\label{app:a2}
\begin{mybox1}
\textbf{English}
\par
\medskip
\textit{1. Ranking order: Four jokers are the highest, followed by any five-of-a-kind bomb, then straight flush, then bombs in descending order of card number (five-of-a-kind, four-of-a-kind), then the seven basic card types (single, pair, triplet, full house, straight, three consecutive pairs).}

\textit{2. Only a full house (three cards of one rank and two cards of another rank) is allowed, not three cards with one, nor four cards with two.}

\textit{3. For bombs, the ranking is based on the number of cards, not the rank of the cards. For example, 3333 $<$  22222.}

\textit{4. The Ace (A) can be the lowest in a straight A2345 or the highest in a straight 10JQKA.}

\textit{5. Four players form two teams, sitting diagonally. The first player to finish is the top winner, followed by the second, third, and bottom. For example, if you finish first, you are the top winner, and your partner is the second winner, which is called a double top. Your opponents are the third and bottom winners, called double bottom. Double top (first winner and second winner) gets a three-level promotion. First winner and third winner get a two-level promotion, and first winner and bottom winner get a one-level promotion.}

\textit{6. The player in the top position gets the highest card and gives tribute. In the next game, the player with the highest tribute starts.}

\textit{7. Declaring cards: When a player has 10 or fewer cards, they must declare their card count once, specifying the exact number.}

\textit{8. Shuffling, cutting, and drawing cards:
In the first game: The player in the east shuffles, the player in the south cuts the deck, and a card is flipped (it must not be a joker or the heart 2). The flip starts from the south, and the person who draws the flipped card starts the game.
In the second game: The upper stream's upstream player shuffles, the upper stream cuts, and the downstream player draws first. In a double bottom scenario, the upper stream's downstream player draws first. The drawing is counterclockwise.}

\medskip

\textbf{Chinese}
\begin{CJK}{UTF8}{gkai}
\par
\medskip

1. 大小：四个王$>$大于五张相同的炸弹$>$同花顺$>$五张相同炸弹$>$四张相同炸弹$>$七种牌型（单张、对子，三同张，三带二、三同连、顺子，三连对）

2. 只能三带二，不能三带一，也不能四带二。

3. 炸弹，比如 3333 $<$ 22222，是按张数，不是按牌面大小。

4. A 在顺子里，可以充当最小的 A2345，也可以充当最大的 10JQKA。

5. 4 个人两两一对，对角坐。第一个出完牌的，是头游，依次是二游，三游，下游。比如，你是头游，你同家的是二游，这叫双上。对手是三游和下游，这叫双下。头游+同家二游（双上），+3 级，也就是原来是 2 级，双上就是 5 级，头游+同家三游，升 2 级，头游+同家下游，升 1 级。

6. 上游拿最大的牌，对应还贡。下一局进贡点大的先出牌。

7. 报牌：手里牌不超过 10 张的情况下必须报牌数，只报一次就行，报具体有几张。

8. 洗、切、抓牌：

首局：东家洗牌，南家切牌，翻一张（不能是王和红桃 2），从南家点数，点到的人就是首抓，抓到翻出的那张是首出牌。

第二局：上游上家洗牌，上游切牌，下游首抓。双下，上游下家首抓。逆时针抓。

\medskip
\end{CJK}

\end{mybox1}

\clearpage

\section{Sample Commentary for Output} \label{app:b}

\subsection{GPT-4}\label{app:b1}
\begin{mybox}
\textbf{English}
\par
\medskip
\textit{1. Lead with the nine of hearts, the opening lead of the nine of hearts is a probing move, aimed at observing opponents' reactions to lower-ranking cards, thus inferring the potential distribution of high-ranking cards and trump cards.}

\textit{2. Follow with the queen of clubs, the player chooses to follow suit with the queen of clubs, indicating a sequence of play and suggesting a coherent hand without the ten or jack of clubs.}

\textit{3. Follow with the king of clubs, continuing to play in sequence indicates that the player's clubs are relatively orderly, without major gaps.}

\textit{4. Pass, the player passing indicates a possible lack of high-ranking cards such as aces or threes, implying a weakness in their hand.}

\textit{5. Follow with the ace of clubs, after assessing the strength of the board, choosing to follow suit with the ace of clubs indicates the player's attempt to control the tempo of the game.}

\textit{6. Follow with the small joker, using the small joker to follow suit is a strong suppressing move, aimed at testing whether the opponent holds the big joker.}

\textit{7. Follow with the big joker, by playing the big joker, the player clearly expresses a strong intent to seize control of the board, while also possibly intending to block certain combinations of cards by opponents.}

\textit{8. Lead with the four of hearts, by playing a low single card, the player may be attempting to clear out scattered cards from their hand, while also possibly hinting at holding higher-ranking cards or trumps.}

\textit{9. Follow with the ten of clubs, this sequence indicates that the player's clubs are relatively orderly, and they are now attempting to build a larger combination.}

\textit{10. Follow with the jack of diamonds, continuing to play smoothly suggests that the player is likely maintaining a balanced hand to facilitate better play in subsequent moves.}

\textit{11. Follow with the queen of spades, this sequence of play continues to demonstrate the coherence of the player's hand.}

\textit{12. Follow with the king of hearts, by playing the king of hearts, the player may be attempting to clear a particular suit, reducing the complexity of their hand.}

\textit{13. Lead with the four of diamonds, playing a low card at this point may be in preparation to clear smaller cards from the hand, maintaining an advantage in high-ranking cards.}

\textit{14. Follow with the nine of diamonds, continuing to play in sequence emphasizes the coherence of the hand and reduces gaps in the board.}

\textit{15. Follow with the three of diamonds, by playing the three of diamonds, the player further confirms the presence of higher-ranking cards and maintains control through low cards.}

\textit{16. Lead with the six of spades, by playing the six of spades, the player attempts to control the board and reduce scattered cards in their hand, maintaining a strategic advantage.}

\textit{17. Follow with the king of spades, by following suit with the king of spades, the player indicates that the previous player may not have larger single cards, continuing to maintain control of the board.}

\textit{18. Follow with the ace of spades, by following suit with the ace of spades, the player may be attempting to further seize control of the board and take the initiative in the game.}

\textit{19. Lead with three nines, by playing three nines, the player may be attempting to deduce the opponents' hand and suppress them through combinations of card types.}

\textit{20. Follow with three queens, continuing to play large cards may be an attempt to build a larger combination, while also trying to suppress opponents.}

\medskip

\end{mybox}

\clearpage 

\begin{mybox}
\textbf{Chinese}
\begin{CJK}{UTF8}{gkai}
\par
\medskip
1. 首出红桃9, 开局首出红桃9是一种试探性的出牌，目的在于观察对手对于较低级牌的反应，从而推断出王牌和高级牌的可能分布。

2. 跟牌梅花Q, 玩家选择跟出梅花Q，这是一种顺牌跟出，表示手中可能没有梅花10或梅花J，暗示了一种较为连贯的牌型。

3. 跟出梅花K, 继续顺序出牌表明玩家手中的梅花较为整齐，没有大的断层。

4. 过牌, 玩家通过过牌表明手中可能缺乏高级牌，如A或3，这暗示了其牌力的一部分弱点。

5. 跟出梅花A,在确认场面上的牌力后，选择跟出梅花A，表明玩家试图控制游戏的节奏。

6. 跟牌小王, 使用小王跟牌，这是一种强力的压制动作，目的在于测试场上是否有大王。

7. 跟牌大王, 通过大王的出牌，玩家明确表达了抢夺牌权的强烈意图，同时也可能意图阻止对手的某些牌型组合。

8. 主动出红桃4, 通过出一个较低的单牌，玩家可能在试图清理手中的散牌，同时也可能暗示手中还持有级牌或王。

9. 跟牌梅花10, 这个顺牌表明玩家手中的梅花较为整齐，此时已经在试图构建更大的牌型。

10. 跟牌方块J, 继续保持牌的顺畅出法，说明玩家可能在保持牌型的平衡，以便于后续更好地出牌。

11. 跟牌黑桃Q, 此时的顺牌出法继续显示出玩家牌型的连贯性。

12. 跟牌红桃K, 通过出红桃K，玩家可能在试图清空某一花色，减少手牌的复杂度。

13. 主动出方块4, 此时出低牌可能是在准备清理小牌，以便手中保持高牌优势。

14. 跟出方块9, 继续顺牌出法，强调牌型的连贯性，减少牌面的断层。

15. 跟出方块3,通过出方块3，玩家进一步验证手中有级牌，并通过低牌维持控制权。

16. 主动出黑桃6, 通过出黑桃6，玩家尝试控制牌权和减少手牌的散牌，维持战略优势。跟出黑桃K,通过跟出黑桃K，玩家显示出上家可能没有更大的单牌，继续维持牌面的控制。

17. 跟出黑桃K,通过跟出黑桃K，玩家显示出上家可能没有更大的单牌，继续维持牌面的控制。

18. 跟牌黑桃A,通过跟牌黑桃A，玩家可能在试图进一步抢夺牌权，掌握牌局的主动权。

19. 主动出三张9,通过出三张9，玩家可能在试图推测对手手中的牌型，并通过牌型组合来压制对手。

20. 跟出三张Q, 继续跟出大牌，可能是在试图构建更大的牌型，同时也可能是在尝试压制对手。
\medskip
\end{CJK}

\end{mybox}

\clearpage 

\subsection{GPT-3.5}\label{app:b2}
\begin{mybox}
\textbf{English}
\par
\medskip
\textit{1. Lead with the nine of hearts: The player starts with the nine of hearts, which is a relatively weak hand, usually used to probe opponents' hands and determine the presence of the big joker.}

\textit{2. Follow with the queen of clubs: Subsequently, the player follows with the queen of clubs, indicating they do not have singles of 10 or jack of clubs.}

\textit{3. Follow with the king of clubs: The player continues to follow with the king of clubs, maintaining the normal sequence of play.}

\textit{4. Pass: The player passes, indicating they may not have aces or threes, and the presence of the big joker is still uncertain.}

\textit{5. Follow with the ace of clubs: The player follows with the ace of clubs, further confirming the presence of the ace of clubs in their hand.}

\textit{6. Press: The player presses, indicating they do not have single-ranking cards.}

\textit{7. Lead with the four of diamonds: The player leads with the four of diamonds, guessing they still have level cards in their hand, possibly more than one.}

\textit{8. Follow with the nine of diamonds: The player follows with the nine of diamonds, indicating they may not have single cards from 5 to 8.}

\textit{9. Pass: The player passes, further confirming the absence of single cards.} 

\textit{10. Follow with the three of diamonds: The player follows with the three of diamonds, further confirming the presence of level cards in their hand, possibly more than one.}

\textit{11. Lead with the six of spades: The player leads with the six of spades, guessing they still have level cards in their hand, aiming to control the board and reduce the number of cards in their hand.}

\textit{12. Follow with the king of spades: The player follows with the king of spades, indicating they may not have single cards, as the highest single card, the king, has already been played.}

\textit{13. Play four jokers with the declaration of ten cards: The player plays four jokers with the declaration of ten cards, indicating they still have multiple level cards in their hand.}

\textit{14. Follow with four jokers: Subsequently, the player follows with four jokers, indicating they are also competing for control of the board.}

\textit{15. Play three sixes with a pair of fours: The player plays three sixes with a pair of fours, indicating they still have two jokers and two other sets of cards.}

\textit{16. Follow with three sevens with a pair of twos: The player follows with three sevens with a pair of twos, forming a suppression.}

\textit{17. Play four twos: The player plays four twos, attempting to charge forward.}

\textit{18. Follow with five fives: Subsequently, the player follows with five fives, indicating they are vying for control.}

\textit{19. Lead with a pair of fours: The player leads with a pair of fours, guessing they are giving cards to the previous player.}

\textit{20. Follow with a pair of eights, declaring eight cards: The player follows with a pair of eights, declaring eight cards.}

\medskip

\end{mybox}

\clearpage 

\begin{mybox}
\textbf{Chinese}
\begin{CJK}{UTF8}{gkai}
\par
\medskip
1. 首出红桃9：玩家以红桃9开始，这是一个比较弱的牌型，通常用于探听对手的牌型和大王的存在。

2. 跟牌梅花Q：后续玩家跟出梅花Q，表明他们没有10或J的单牌。

3. 跟出梅花K：玩家继续跟出梅花K，保持正常的顺牌。

4. 过牌：玩家过牌，表明他们可能没有A和3，而大王的存在还不能确定。

5. 跟出梅花A：玩家跟出梅花A，进一步证实了他们手中有梅花A。

6. 压牌：玩家压牌，表明他们手中没有单级牌。

7. 主动出方块4：玩家主动出方块4，猜测手中还有级牌，并且不止一张。

8. 跟出方块9：玩家跟出方块9，表明他们可能没有5到8的单牌。

9. 过牌：玩家过牌，进一步验证没有单牌。

10. 跟出方块3：玩家跟出方块3，进一步证实了手中有级牌，并且不止一张。

11. 主动出黑桃6：玩家主动出黑桃6，猜测手中还有级牌，目的是控制牌权和减少手牌套数。

12. 跟出黑桃K：玩家跟出黑桃K，表明他们手中可能没有单牌，因为最大的单牌已经出到了K。

13. 4张10炸弹，报牌10张：玩家出4张10的炸弹，并且报牌10张，表明他们手中还有多张级牌。

14. 跟出4张J炸弹：后续玩家跟出4张J的炸弹，表明他们也在争夺牌权。

15. 出牌3张6带对4：玩家出3张6带对4，表明他们手中还有2个炸弹和2套其他牌型。

16. 跟牌3张7带对2：玩家跟牌3张7带对2，形成压制。

17. 4张2炸弹：玩家出4张2的炸弹，尝试冲锋。

18. 跟牌5张5：后续玩家跟牌5张5，表明他们在抢夺牌权。

19. 主动出对4：玩家出对4，猜测是给上家送牌。

20. 跟牌对8，报牌8张：玩家跟牌对8，并且报牌8。
\medskip
\end{CJK}

\end{mybox}

\clearpage 

\subsection{Qwen-32B}\label{app:b3}
\begin{mybox}
\textbf{English}
\par
\medskip
\textit{1. Firstly, playing the 9 of hearts is a common opening strategy to test the distribution of the opponents' level cards and kings.}

\textit{2. Opponent calls the queen of clubs, which is a normal straight action, and could mean that he doesn't have the 10 and the jack singleton, or that they have been used as part of a bomb.}

\textit{3. Following with the king of clubs also fits the sequence logic,  indicating that the 10 and Jack of clubs are either retained or no longer in their hand.}

\textit{4. Passing the turn, possibly due to having no better card in hand or wanting to observe the situation.}

\textit{5. Passing again, as the teammate's cards do not significantly impact the current decision-making, hence no need to focus on them.}

\textit{6. Following with the ace of clubs means either there is an A-K pair or just a single ace, also attempting to take control of the hand.}

\textit{7. Continuing to pass, possibly because there are no single level cards, or the single level cards are already part of other hands.}

\textit{8. Still passing, as the teammate’s hand does not affect one's own strategy.}

\textit{9. Following with the small joker might be because the player holds the big joker, aiming to control the situation.}

\textit{10. Passing, with no indication of the big joker.}

\textit{11. Following with the big joker, clearly to suppress the opponents and vie for control of the hand.}

\textit{12. Continuing to pass, because it is believed there are no better opportunities to play.}

\textit{13. Leading with the 4 of hearts, because there are level cards or a king in hand, attempting to take the initiative.}

\textit{14. Following with the 10 of clubs indicates there might be a 10-K pair in hand, or all single cards below 10 have been played.}

\textit{15. Passing, possibly because there are no single cards bigger than the 10 of clubs.}

\textit{16. Playing the Jack of diamonds follows the rule of sequence.}

\textit{17. Playing the queen of spades also maintains the sequential rhythm.}

\textit{18. Following with the king of hearts indicates there might be a K-Q pair or higher single cards in hand.}

\textit{19. Continuing to pass, because it is thought there are no better opportunities to play.}

\textit{20.Playing the big joker is an obvious suppression move, attempting to control the hand.}

\medskip

\end{mybox}

\clearpage 

\begin{mybox}
\textbf{Chinese}
\begin{CJK}{UTF8}{gkai}
\par
\medskip
1. 首先，红桃9的出牌是为了试探对方的级牌和王的分布，这是一种常见的开局策略。

2. 对方跟牌梅花Q，这是一个正常的顺牌动作，可能意味着他没有10和J的单牌，或者它们已经被用作炸弹的一部分。

3. 梅花K的跟进同样符合顺牌逻辑，表明梅花10和J已被保留或已经不在他们的手中。

4. 过牌，这是由于手中可能没有更好的牌或者想要观察局势。

5. 再次过牌，队友的牌对于当前决策影响不大，因此无需关注。

6. 梅花A的跟进意味着要么有A-K连对，要么就是单A，同时也在尝试控制牌权。

7. 继续过牌，可能是因为没有单级牌，或者单级牌已经构成其他牌型。

8. 依旧过牌，队友的牌型不影响自己的策略。

9. 小王的跟牌可能是因为手中有大王，为了控制局势。

10. 过牌，没有大王的信号。

11. 大王的跟进，显然是为了压制对手并争取牌权。

12. 连续过牌，因为认为没有更好的出牌机会。

13. 红桃4的主动出牌，因为手里有单级牌，或者王，尝试获取主动权。

14. 梅花10的跟牌，表明手中可能有10-K的连对，或者10以下的单牌都已经出完。

15. 过牌，可能是没有比梅花10更大的单牌。

16. 方块J的出牌，遵循了顺牌的规律。

17. 黑桃Q的出牌，同样保持了顺牌的节奏。

18. 红桃K的跟牌，表示手中有K-Q连对或者更高单牌。

19. 连续过牌，因为认为没有更好的出牌时机。

20. 大王的出牌，是明显的压制行动，试图控制牌权。
\medskip
\end{CJK}

\end{mybox}

\clearpage 

\subsection{Yi-34B}\label{app:b4}
\begin{mybox}
\textbf{English}
\par
\medskip
\textit{1. Player A leads with the 9 of Hearts, typically a probing play to gauge whether opponents hold high cards or specific combinations like sequences or pairs.}

\textit{2. Player B follows with the Queen of Clubs, suggesting B might have a sequence or is testing A's hand.}

\textit{3. Player C plays the King of Clubs, possibly to maintain a sequence or guess the hands of A and B.}

\textit{4. Player D passes, likely indicating no suitable cards to play or waiting for a better opportunity.}

\textit{5. Player A passes again, possibly to keep high cards in hand or waiting for the right moment.}

\textit{6. Player B plays the Ace of Clubs, possibly to maintain a sequence or demonstrate strength.}

\textit{7. Player C passes again, perhaps to conserve strong cards or waiting for an opportune moment.}

\textit{8. Player D passes again, likely to hold onto high cards or wait for a better time.}

\textit{9. Player A plays the Small Joker, possibly to maintain a sequence or show strength.}

\textit{10. Player B passes again, possibly conserving strong cards or waiting for the right moment.}

\textit{11. Player C passes again, likely keeping strong cards or waiting for an opportune time.}

\textit{12. Player D plays the Big Joker, possibly to maintain a sequence or demonstrate strength.}

\textit{13. Player A overcalls, indicating confidence that their hand is stronger than the others'.}

\textit{14. Player B passes again, possibly to hold onto strong cards or wait for the right moment.}

\textit{15. Player C passes again, likely keeping strong cards or waiting for an opportune time.}

\textit{16. Player D overcalls, showing confidence that their hand is stronger than the others'.}

\textit{17.layer A passes again, possibly to keep strong cards in hand or waiting for the right moment.}

\textit{18. Player B passes again, possibly conserving strong cards or waiting for the right moment.}

\textit{19. Player C passes again, likely keeping strong cards or waiting for an opportune time.}

\textit{20. Player D leads with the 4 of Hearts, possibly to test if other players have a sequence or pair of Hearts.}

\medskip

\end{mybox}

\clearpage 

\begin{mybox}
\textbf{Chinese}
\begin{CJK}{UTF8}{gkai}
\par
\medskip
1. 首出红桃9：玩家A首出红桃9，这通常是试探性的出牌，旨在了解对手手中是否有大牌或者特定的牌型，比如顺子、对子等。

2. 跟牌梅花Q：玩家B跟牌梅花Q，这表明B可能有一手顺子或者是在试探A的牌型。

3. 跟出梅花K：玩家C跟出梅花K，这可能是为了保持顺子或者是在猜测A和B的牌

4. 过牌：玩家D选择过牌，这可能意味着D没有合适的牌可以跟出，或者是在等待更好的出牌机会。

5. 过牌：玩家A再次过牌，这可能是为了保留手中的大牌或者是在等待合适的时机。

6. 跟出梅花A：玩家B跟出梅花A，这可能是为了保持顺子或者是在显示自己的实力。

7. 过牌：玩家C再次过牌，这可能是为了保留手中的大牌或者是在等待合适的时机。

8. 过牌：玩家D再次过牌，这可能是为了保留手中的大牌或者是在等待合适的时机。

9. 跟牌小王：玩家A跟牌小王，这可能是为了保持顺子或者是在显示自己的实力。

10. 过牌：玩家B再次过牌，这可能是为了保留手中的大牌或者是在等待合适的时机。

11. 过牌：玩家C再次过牌，这可能是为了保留手中的大牌或者是在等待合适的时机。

12. 跟牌大王：玩家D跟牌大王，这可能是为了保持顺子或者是在显示自己的实力。

13. 压牌：玩家A压牌，这表明A有信心自己的牌型比其他玩家都大。

14. 过牌：玩家B再次过牌，这可能是为了保留手中的大牌或者是在等待合适的时机。

15. 过牌：玩家C再次过牌，这可能是为了保留手中的大牌或者是在等待合适的时机。

16.压牌：玩家D压牌，这表明D有信心自己的牌型比其他玩家都大。

17.过牌：玩家A再次过牌，这可能是为了保留手中的大牌或者是在等待合适的时机。

18. 过牌：玩家B再次过牌，这可能是为了保留手中的大牌或者是在等待合适的时机。

19. 过牌：玩家C再次过牌，这可能是为了保留手中的大牌或者是在等待合适的时机。

20. 主动出红桃4：玩家D主动出红桃4，这可能是为了试探其他玩家是否有红桃的顺子或者是对子。
\medskip
\end{CJK}

\end{mybox}

\clearpage

\subsection{GLM-4}\label{app:b5}
\begin{mybox}
\textbf{English}
\par
\medskip
\textit{1. Leading with the 9 of Hearts, testing the distribution of trump cards and kings.}

\textit{2. Following suit with the Queen of Clubs, a normal sequence play, likely not holding singleton 10 or Jack.}

\textit{3. Following with the King of Clubs, a natural sequential play.}

\textit{4. Passing, possible absence of the Ace and 3, high likelihood of not having a singleton Ace, kings are uncertain.}

\textit{5. Passing, unnecessary for teammate to take action.}

\textit{6. Following with the Ace of Clubs, guessing no singleton trump cards, trumps may form other combinations or no trumps at all, kings remain uncertain.}

\textit{7.Passing, unnecessary for teammate to take action.}

\textit{8. Following with the Small Joker, guessing no Big Joker.}

\textit{9. Playing the Big Joker, asserting, seizing control.}

\textit{10. Leading with the 4 of Hearts, guessing holding either a singleton trump or kings.}

\textit{11. Following with the 10 of Clubs, guessing no cards larger than 10, uncertain if there's a singleton trump or king.}

\textit{12. Following with the Jack of Diamonds, a normal sequential play.}

\textit{13. Following with the Queen of Spades, a natural sequence play.}

\textit{14. Following with the King of Hearts, a normal sequence play.}

\textit{15. Passing, no need for teammate to take action.}

\textit{16. Over-calling, no other choice. It can be inferred that there might not be a singleton trump card.}

\textit{17. Passing, unnecessary for teammate to take action.}

\textit{18. Over-calling, seizing control of the round. Currently, the highest card in play is the trump 3, confirming neither opponent nor partner have the singleton trump 3, and since I lack trumps, there’s a higher likelihood that an opponent still has it.}

\textit{19. Leading with the 4 of Diamonds, guessing holding more than one trump card in hand.}

\textit{20. Followed by the 9 of diamonds, a normal straight, and one can guess that there is no singleton from 5 to 8.}

\medskip

\end{mybox}

\clearpage 

\begin{mybox}
\textbf{Chinese}
\begin{CJK}{UTF8}{gkai}
\par
\medskip
1. 首出红桃9，试探级牌和王牌的分布。

2. 跟牌梅花Q，正常顺牌，可能没有10、J单牌。

3. 跟出梅花K，正常顺牌。

4. 过牌，可能没有A和3，大概率没有单A，王无法确认。

5. 过牌，队友的牌没有必要管。

6. 跟出梅花A，猜测没有单牌的级牌，级牌组成了其他牌型或没有级牌，王无法确认。

7. 过牌，队友的牌没有必要管。

8. 跟牌小王，猜测没有大王。

9. 跟牌大王，压制，抢夺牌权。

10. 主动出红桃4，猜测手中有单级牌或王。

11. 跟牌梅花10，猜测没有大于10的单牌，无法确定是否有单级牌或王。

12. 跟牌方块J，正常顺牌。

13. 跟牌黑桃Q，正常顺牌。

14. 跟牌红桃K，正常顺牌。

15. 过牌，队友的牌没必要管。

16.压牌，别无选择。可以推测手中没有单级牌。

17.过牌，队友的牌没有必要管。

18. 压牌，抢夺牌权。目前场面最大的是级牌3，可以确认的是，对家和上家都没有单级牌3，而我自己没有级牌，那么级牌3出现炸弹的可能性较高。

19. 主动出方块4，猜测手中还有级牌，且不止1张。

20. 跟出方块9，正常顺牌，可以猜测没有5-8的单牌。
\medskip
\end{CJK}

\end{mybox}

\clearpage

\subsection{Our}\label{app:b6}
\begin{mybox}
\textbf{English}
\par
\medskip
\textit{1. Pay attention to the card-playing patterns of the players to your right and left, especially whether they tend to play pairs or sequences. This could help you infer the possible card combinations they might have. Don't rush to play high single cards, particularly when you're unsure if your opponents hold them. Often, the last few cards are the most crucial as they might include key combinations that determine the outcome. Given the current scenario and your cards, I suggest initially testing your opponents' hands. You might consider playing some small pairs (like 3, 4) or medium-sized singles (like the 7 of hearts). This way, even if the opponents have high cards, they might pause in front of your small pair. Then, based on their responses, gradually increase the strength and value of your cards until the final sprint.}

\textit{2. Now, there are still three 2s unplayed, and our opposite player might have strong cards. If they do, we could try playing some high cards to extend our advantage. However, if they don't have strong cards, we need to be more cautious with our plays, ensuring each card maximizes our scoring potential while closely monitoring our opponents' moves to adjust our strategy accordingly.}

\textit{3. Because there are still likely 2s and Aces that can change the situation on the table. After several rounds of passing, the players found that their opponents weren't eager to play high cards, suggesting a weak hand. The player cautiously chose to play a cross shape with a pair of Kings, hoping for a response from the following player. However, the next player continued to pass. After observing the previous player play three 3s with a heart, the player decided to challenge with a combination of four Queens to pressure the opponents and to gauge their probable hand types. The opposite player chose not to 'fire back,' further confirming the earlier speculation about their weak cards.}

\textit{4. You played the 10 of hearts, indicating you might have higher cards. The opponents responded with a pair of Aces, showing they were also actively aiming for victory. A 9 of spades was then played, to which you responded with a Jack of spades, indicating your strong hand. They quickly followed with a Queen of hearts, applying more pressure. You countered with a King of hearts, displaying confidence in this round. Subsequently, they played a pair consisting of the 2 of diamonds and the 3 of clubs to expand their lead, but you suppressed them with your strong Ace of spades. Unyielding, they attempted to find an opportunity with a 6 of clubs, but you remained composed, deciding not to play yet, waiting for a better moment to continue the game.}

\textit{5. The previous player started with a 4 of hearts, and the next played the Ace of clubs. From my side, I played the 7, Queen, and Ace of spades, attempting to regain control, but the opposite player responded with the 10, Jack, Queen, and Ace of diamonds, showing they were strong contenders. The previous player then laid out a sequence of hearts: 2, 5, 6, 7, and King, indicating they might have a flush or straight. The next player followed with a sequence of clubs: 3, 4, 8, 9, and Jack, suggesting they might also hold strong clubs. I played the 2 and 4 of spades, trying to confuse the situation and defend with the 10 or Queen of clubs if necessary.}

\textit{6. In this game, the player observed their hand containing several pairs, such as the 2 and 4 of diamonds, a pair of 5s, and two Kings, along with scattered cards like the 8 of clubs, King, various hearts, and the 4, 5, 8 of spades. In the first round, other players passed without taking the lead. In the second round of playing, the next player chose a pair of Jacks, and after considering, the player decided to follow with their single King of hearts.}

\textit{7. The previous player played a pair of diamond 3s, and you chose to pass. The player opposite also passed. The next player played a single spade Ace, and you still passed. The opposite player then played a pair of heart 10s, and you chose to suppress it with a small Joker, due to the lack of direct combating combinations in your hand. From my current hand, I need to form pairs, sequences, or flushes to counter the opponents.}

\textit{8.Currently, our hand includes a diamond 2 with an extra card, pairs of 7s and Kings, with single cards of club 8, club King, and heart 2. There's still one 2 left outside, three Aces, and some cards that might form a sequence or flush. The previous player puts out a pair of 7s which we still can't follow, waiting for an opportunity or to disrupt the card types. Next, we need to observe the situation on the table, looking for the right opportunity to make a move.}

\textit{9. In this round of Mahjong, the players' hands are highly diverse. They hold a diamond sequence from 2 to Ace, with two Kings, a pair of 7s, and a pair of Queens. The clubs also have several consecutive pairs and single cards like 8, King, 2, 5, 6 (duplicate), 7, and Queen. Hearts include 2, 3, 4, 5, 6, 7, duplicated 10s, Jack, and two Kings. Lastly, there are spades 4, duplicated 5s, 8, 9, Jack, King, and two Aces. The previous player continues to play single cards diamond 4 and diamond 9, but the player does not have the appropriate cards to follow. As the current hand cannot form effective sequences or pairs, the player opts for a passing strategy, waiting for a better chance to make a move. This series of actions shows the patience and tactical considerations in the Mahjong game, not rushing for immediate gains but aiming for steady victory.}

\textit{10. My hand still holds a diamond Ace, club Queen, club King, and spade Queen, spade King. I decide to play using a three with a pair approach: club Queen, club King, spade Queen, spade King, with the diamond Ace as a single card. This method in Guandan usually effectively clears the field and applies pressure on the opponents. Considering the possible counter strategies of the opponents, they might choose to follow or pass. If the opponents have the corresponding pairs or consecutive flushes, they might follow; if not, they might choose to pass, waiting for a better opportunity to play.}

\medskip

\end{mybox}

\clearpage 

\begin{mybox}
\textbf{Chinese}
\begin{CJK}{UTF8}{gkai}
\par
\medskip
1. 留意对家和下家的出牌模式，特别是他们是否倾向于出对子或者顺子，这将帮助你推测他们手中可能有的牌型。  不要急于出单张的大牌，特别是当你不确定对手是否持有这些牌时。通常情况下，最后的几张牌是最重要的，因为它们可能包含决定胜负的关键组合。 根据当前局势和手牌，我建议先试探下家和对家的手牌情况。你可以考虑先出一些小对子（如3、4）或者中等大小的单张（如红桃7）。这样即使他们有大牌，也可能在你的小对子面前暂时停下。 之后，根据对手的反馈，逐步提高你出牌的强度和价值，直到最后的冲刺阶段。

2. 现在外面还有三张2，我们对家可能有火。如果对家有火，我们可以尝试打一些火来扩大优势。但如果对家没有火，我们需要更加谨慎地出牌，确保每一张牌都能最大化我们的得分。同时，也要密切关注对手的动向，及时调整策略应对。

3. 因为外面还有可能的2和A来改变局势。经过几轮的过牌，玩家发现对手并未急于出火，推测他们手上的牌可能并不支持。于是玩家谨慎地选择打出了一个十字形带对K，希望下家能给出回应。然而下家仍然没有配合，继续过牌。  在观察到上家出了三张3配红桃后，玩家决定挑战一下，打出4个Q的组合来给对手压力，并借此机会判断他们手上可能的牌型。结果对家并没有选择开火，这进一步确认了之前关于火牌的推测。

4. 你打出红桃10，显示出你的手牌中可能有较大的牌面。对手的应对是一对A，他们似乎也在积极争取这局的胜利。紧接着，黑桃9加入战局，你用黑桃J进行回应，显示你的牌力不容小觑。对方紧接着亮出红桃Q，再次加压。你以红桃K进行回击，展现你对这一轮对决的信心。然后，对方打出方块2和梅花3的对子，试图扩大领先优势，但你用强大的黑桃A压制了他们。对手不屈不挠，试图通过一张梅花6寻找机会，但你保持冷静，决定暂时不出手，等待更好的时机让牌局继续进行。

5. 上家先手出了红桃4，下家则跟了梅花A。我这边出黑桃7，黑桃Q，黑桃A，试图夺回控牌权，但对家用方块10，方块J，方块Q，方块A回应，表明他有足够的实力进行对抗。上家随后打出红桃的连牌，红桃2，红桃5，红桃6，红桃7和红桃K，显示他的手牌中可能有同花或顺子，下家则跟进梅花3，梅花4，梅花8，梅花9，梅花J，这说明他手中可能也有梅花的大牌。我出黑桃2，黑桃4，试图混淆局势，并利用梅花10或梅花Q进行防御。

6. 这一局游戏里，玩家观察到手上的牌情，包括了多对子，如方块2和4，还有对5，以及两个K，同时还有散牌梅花8，K，红桃系列以及黑桃4，5，8等。首局其他玩家过牌，没有主动出击。第二轮出牌时，下家选择了一个对J的牌型进行出牌，此时玩家考量后决定用自己手里的单张红桃K跟注。

7. 上家打出了方块3对子，你选择过牌。 对家也选择过牌。下家打出单张黑桃A，你依然过牌。 对家这次出了一个红桃10对子，你选择用小王进行压制。是因为你的手牌里缺乏可以直接对抗的组合。从我当前的手牌来看，我需要构建出对子、顺子或者同花等有利组合来对抗对手

8. 现在我们手上有一个方块的二带一，对子有对七和一对K，单张有梅花8、梅花K和红桃2。目前外面还剩一张二，三张A，以及一些可能构成顺子或同花的牌。上家出对7，我们还是无法接，等待机会或者打散牌型。接下来需要观察场上局势，寻找合适的机会出手。

9. 在这一轮的麻将游戏中，玩家手上的牌型非常多样化。他们拥有方块的顺子从2到A，并且有两个K，一对7和一对Q，梅花也有不少连对和单张，如8、K、2、5、6（重复）、7和Q。红桃则包括了2、3、4、5、6、7、10（重复）以及J和两个K。最后还有黑桃的4、5（重复）、8、9、J、K和2张A。  上家连续出单牌方块4和方块9，但玩家的手中并没有合适的牌来进行跟打。由于当前手牌组合无法组成有效的顺子或对子，玩家选择了过牌策略，等待更好的出手机会。这一系列操作显示了玩家在麻将游戏中的耐心和战术考量，不急于一时之得失，而是稳中求胜。

10. 我的手牌中还剩下方块A、梅花Q、K和黑桃Q、K。我决定使用三带二的方式出牌：梅花Q、K，黑桃Q、K，外加方块A作为单张。这种出法在掼蛋中通常能够有效地清场并给对手施加压力。  考虑到对手可能的应对策略，他们可能会选择跟牌或过牌。如果对手有相应的对子或同花连对，他们可能会跟进；如果没有，他们可能会选择过牌，等待更好的出手时机。

\medskip
\end{CJK}

\end{mybox}


\end{document}